%% file: ijcai19.tex
\documentclass{article}
\usepackage{ijcai19}

\usepackage{times}
\usepackage{soul}
\usepackage{url}
\usepackage[hidelinks]{hyperref}
\usepackage[utf8]{inputenc}
\usepackage[small]{caption}
\usepackage{graphicx}
\usepackage{amsmath}
\usepackage{booktabs}
\usepackage{algorithm}
\usepackage{algorithmic}
\usepackage{comment}

\usepackage{enumitem}
\input{math_commands.tex}

\usepackage{subcaption}

\title{ARMIN: Towards a More Efficient and Light-weight Recurrent Memory Network}
\author{
Zhangheng Li$^{1,2}$\and
Jia-Xing Zhong$^{1,2}$\and
Jingjia Huang$^{1,2}$\and
Tao Zhang$^{1,2}$\and
Thomas Li$^1$\And
Ge Li$^{1,2}$\footnote{Contact Author}\\
\affiliations
$^1$School of Electronic and Computer Engineering, Peking University\\
$^2$Peng Cheng Laboratory\\
\emails
\{zhanghengli, jxzhong, jjhuang, t\_zhang\}@pku.edu.cn,
Thomasli@pkusz.edu.cn,
geli@ece.pku.edu.cn
}

\begin{document}

\maketitle

\begin{abstract}
 In recent years, memory-augmented neural networks(MANNs) have shown promising power to enhance the memory ability of neural networks for sequential processing tasks. However, previous MANNs suffer from complex memory addressing mechanism, making them relatively hard to train and causing computational overheads. Moreover, many of them reuse the classical RNN structure such as LSTM for memory processing, causing inefficient exploitations of memory information. In this paper, we introduce a novel MANN, the Auto-addressing and Recurrent Memory Integrating Network (ARMIN) to address these issues. The ARMIN only utilizes hidden state $h_t$ for automatic memory addressing, and uses a novel RNN cell for refined integration of memory information. Empirical results on a variety of experiments demonstrate that the ARMIN is more light-weight and efficient compared to existing memory networks. Moreover, we demonstrate that the ARMIN can achieve much lower computational overhead than vanilla LSTM while keeping similar performances. Codes are available on \href{https://github.com/zoharli/armin}{github.com/zoharli/armin.}
\end{abstract}

\section{Introduction}
Recurrent neural networks, such as the Long Short-Term Memory (LSTM) \cite{hochreiter1997long} and Gated Recurrent Unit (GRU) \cite{cho2014learning} have shown good performance for processing sequential data. However, it's known that RNNs suffer from gradient vanishing problem. Moreover, as pointed out by Rae $et\ al.$ \shortcite{rae2016scaling}, the number of parameters grows proportionally to the square of the size of the hidden units, which carry the historical information. Recently, memory-augmented neural networks exhibit promising power to address these issues, by decoupling memory capacity from model parameters, $i.e.$ maintaining an external memory, and backpropagating the gradients through the memory.

Neural Turing Machine (NTM) \cite{graves2014neural} first emerged as a recurrent model that incorporates external memory abilities. NTM maintains a memory matrix, and at every time-step, the network reads and writes (with erasing) to the memory matrix using certain soft-attentional mechanism, controlled by an LSTM that produces read and write vectors. NTM and its successor, the Differentiable Neural Computer\cite{graves2016hybrid}, have shown success on some algorithmic tasks such as copying, priority sorting and some real-world tasks such as question answering. But one limitation of the NTM is that due to its smooth read and write mechanism, NTM has to do propagations on the entire memory, which is not neccessary and may cause high computational overhead when the memory is in large-scale. However, these two external memory models have relatively complicated and hand-crafted memory addressing mechanisms, making the backpropagations through memory not straightforward and also causing high computational overheads. Moreover, they basically reuse the classical RNN structure such as LSTM for memory processing, which causes inefficient exploitations of memory information. The RNN in these models plays a simple role of being a controller, but the gradient vanishing problem of RNN itself is not given enough attention, which usually causes a degradation in the training speed and final performance.

 Inspired by prior memory models, progresses have been made to build a bridge between simple RNNs and complicated memory models. Kurach $et\ al.$ \shortcite{kurach2015neural} propose the Neural Random-access Machines (NRAM) that can manipulate and dereference pointers to an external variable-size random-access memory. Danihelka $et\ al.$ \shortcite{danihelka2016associative} improve LSTM with ideas from holographic reduced representations that enables key-value storage of data. Grave $et\ al.$ \shortcite{grave2016improving} propose a method of augmenting LSTM by storing previous (hidden state, input word) pairs in memory and using the current hidden state as a query vector to recover historical input words. This method requires no backpropagation through memory and performs well on word-level language tasks.Grefenstette \shortcite{grefenstette2015learning}, Dyer \shortcite{dyer2015transition}, Joulin \shortcite{joulin2015inferring} augment RNNs with a stack structure that works as a natural parsing tool, and use them to process algorithmic and nature language processing (NLP) tasks; nonetheless, the running speed of stack-augmented RNNs is rather slow due to multiple push-pop operations at every time-step. Rae $et\ al.$ \shortcite{rae2016scaling} proposes the Sparse Access Memory(SAM) network, by thresholding memory modifications to a sparse subset and using the approximate nearest neighbor (ANN) index to query the memory. Ke $et\ al.$ \shortcite{ke2018sparse} propose the Sparse Attentive Backtracking (SAB) architecture, which recalls a few past hidden sates at every time-step and do ``mental" backpropagations to the nearby hidden states with respect to the recalled hidden states. Gulcehre $et\ al.$ \shortcite{gulcehre2017memory} propose the TARDIS network, which recalls a single memory entry at each time-step and use an LSTM-resembled RNN to process memory information. However, the TARDIS still involves some hand-crafted memory addressing methods which cause a considerable amount of computational overhead.
 
Based on the motivation of proposing a more efficient, light-weight and universal method of combining RNNs and MANNs, we introduce the ARMIN architecture, a recurrent MANN with a simple and straightforward memory addressing mechanism and a novel RNN cell for refined memory information processing. Concretely, our contributions are as follows:

\begin{itemize}
	\item We propose a simple yet effective memory addressing mechanism for the external memory, namely Auto-addressing, by encoding the information for memory addressing directly via the inputs $\vx_t$ and the hidden states $\vh_{t-1}$.
	\item We propose a novel RNN cell that allows refined control of information flow and integration between memory and RNN structure. With only a single such cell, it achieves better performance than many hierarchical RNN structures in our character-level language modelling task.
	\item We show that the ARMIN is robust to small iteration lengths when training long sequential data, which enables training with large batch sizes and boost the training speed. 
	\item We demonstrate competitive results on various tasks while keeping efficient time and memory consumption during training and inference time.
\end{itemize}

\section{Background}

\subsection{Gumbel-softmax Estimator}
\label{gumbel}

Categorical distribution is a natural choice for representing discrete structure in the world. However, it's rarely used in neural networks due to its inability to backpropagate through samples \cite{jang2016categorical}. To this end, Maddison \shortcite{maddison2016concrete} and Jang \shortcite{jang2016categorical} propose a continuous relaxation of categorical distribution and the corresponding gumbel-softmax gradient estimator that replaces the non-differentiable sample from a categorical distribution with a differentiable sample. Specifically, given a probability distribution $\vp=(\pi_1,\pi_2,...,\pi_k)$ over $k$ categories, the gumbel-softmax estimator produces an one-hot sample vector $\vy$ with its $i$-th element calculated as follows:
\begin{equation}
y_i=\frac{\exp((\log (\pi_i)+g_i)/\tau)}{\sum_{j=1}^k\exp((\log(\pi_j)+g_j)/\tau)}\ \ \ \text{for}\ i=1,2,...,k,
\end{equation}
where $g_1,...,g_k$ are i.i.d samples drawn from Gumbel distribution \cite{gumbel1954statistical}: 
\begin{equation}
g_i=-\log(-\log(u_i))\ , u_i\sim \text{Uniform}(0,1),
\end{equation}
and $\tau$ is the $temperature$ parameter. In practice, we usually start at a high temperature and anneal to a small but non-zero temperature \cite{jang2016categorical}. The gumbel-softmax function have been used in many recurrent networks such as TARDIS \cite{gulcehre2017memory}, and Gumbel-Gate LSTM \cite{li2018towards}.

\section{Auto-addressing and Recurrent Memory Integrating Network}
\label{headings}

\begin{figure}[tb]
\begin{center}
\includegraphics[width=\linewidth]{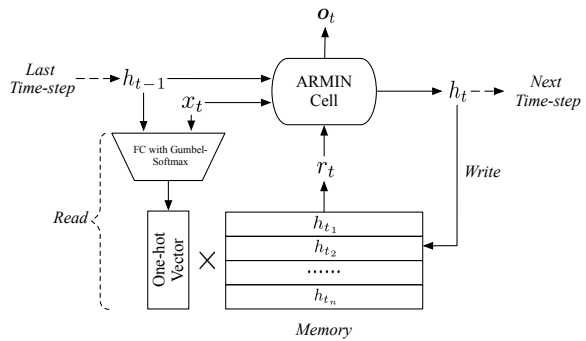}
\end{center}
\vspace{-17\baselineskip}
\caption{The ARMIN structure. At each time-step, the ARMIN performs read operation, cell processing and write operation in chronological order: {\bf (a)} It reads out a historical hidden state $\vr_t$ from memory with an one-hot read vector produced via passing $\vx_t$ and $\vh_{t-1}$ to a fully connected layer followed by a gumbel-softmax function. {\bf (b)} The ARMIN cell receives $\vx_t$, $\vh_{t-1}$ and $\vr_t$ as inputs and outputs $\vo_t$ and $\vh_t$. $\vo_t$ is passed to output layers, and $\vh_t$ is passed to next time-step. {\bf (c)} $\vh_t$ is written to the previous location of $\vr_t$.}
\label{ARMIN figure}
\end{figure}

In this section, we describe the structure of the ARMIN network as shown in Figure~\ref{ARMIN figure}. It consists of a recurrent cell and an external memory that stores historical hidden states. While processing sequential data, the ARMIN performs reading from memory, cell processing and writing to memory operations in chronological order during each time-step. In the following subsections, we first explain the structure of the recurrent cell, and then discuss the read and write operations.

\subsection{The Recurrent Cell of ARMIN}

By combining the gating mechanism of classical LSTM structure, we propose a novel recurrent cell structure for memory information processing, namely the ARMIN cell. At every time-step, it takes in an input $\displaystyle \vx_t$, the last hidden state $\displaystyle \vh_{t-1}$ and a recovered historical hidden state $\displaystyle \vr_t$ chosen by a read operation, and produces an output vector $\displaystyle \vo_t$ and the new hidden state $\displaystyle \vh_t$. The computation process is as follows:

\begin{equation}
\label{eq8}
\left\{
\begin{aligned}
\vg^h_t \\
\vg^r_t
\end{aligned}
\right\}=
\left\{
\begin{aligned}
\sigma \\
\sigma
\end{aligned}
\right\}\mW_{ig}[\ \vx_t,\vh_{t-1},\vr_t\ ]+\vb_{ig}\ ,
\end{equation}
\begin{equation}
\label{eq9}
\vh^g_{t-1}=\vg^h_t \circ \vh_{t-1}\ ,
\end{equation}
\begin{equation}
\label{eq10}
\vr^g_{t}=\vg^r_t \circ \vr_{t}\ ,
\end{equation}
\begin{equation}
\label{eq11}
\left\{
\begin{aligned}
\vi_t\  \\ \vf_t\  \\ \vg_t\  \\ \vo^h_t \\ \vo^r_t 
\end{aligned}
\right\}=
\left\{
\begin{aligned}
\sigma\ \   \\ \sigma\ \   \\ \tanh \\ \sigma\ \  \\ \sigma\ \ 
\end{aligned}
\right\}\mW_{go}[\ \vx_t,\vh^g_{t-1},\vr^g_t\ ]+\vb_{go}\ ,
\end{equation}
\begin{equation}
\label{eq12}
\vh_{t}=\vf_t \circ \vh_{t-1}\ + \vi_t\  \circ \vg_t\ ,
\end{equation}
\begin{equation}
\label{eq13}
\vo_{t}=[\ \vo^h_t \circ \tanh(\vh_{t})\ ,\vo^r_t \circ \tanh(\vr_{t})\ ]\ .
\end{equation}

where  $\textstyle \vh_{t-1},\vh_t \in \R^{d_h}, \vr_t \in \R^{d_r}, \vx_t \in \R^{d_i}$, and $\mW_{ig} \in \R^{2d_h \times (d_i+d_h+d_r)}, \mW_{go} \in \R^{(4d_h+d_r)\times (d_i+d_h+d_r)}, \vb_{ig} \in \R^{d_h+d_r}, \vb_{go} \in \R^{4d_h+d_r}, \vo_t \in \R^{d_h+d_r}$. We refer to $d_h$ as the $hidden\ size$ of the recurrent cell of ARMIN. Usually we have $d_h=d_r$ and allocate equal number of weight parameters for $\vh_t$ and $\vr_t$.

In equation \ref{eq8} $\sim$ \ref{eq10}, two gates are calculated to control the information flow for $\vh_{t-1}$ and $\vr_t$ respectively, generating gated hidden state $\vh^g_{t-1}$ and historical state $\vr^g_{t}$.  Then as shown in equation \ref{eq11}, we compute the input gate $\vi_t$, forget gate $\vf_t$, cell state $\vg_t$ and output gate $\vo^h_t$ for the new hidden state just like in classical LSTM structure. Additionally, an output gate $\vo^r_t$ for historical state $\vr_t$ is computed. Next in equation \ref{eq12}, we compute new hidden state $\vh_t$ that is the sum of $\vh_{t-1}$ and cell state $\vg_t$, leveraged by forget gate $\vf_t$ and input gate $\vi_t$. Finally in equation \ref{eq13}, we calculate the output of this time-step, which is the concatenation of the gated contents from $\vh_t$ and $\vr_t$. 
	
Intuitively, the $\vh_{t-1}$ acts as the old working memory, and $\vr_t$ is treated as the long-term memory. The cell processes them with the input $\vx_t$ to generate the new working memory $\vh_t$ and the output $\vo_t$. More specifically, each $\vr_t$ is a summary of historical hidden states selected by auto-addressing mechanism. The ARMIN cell learns to recurrently integrate the summary of long-term information from $\vr_t$ into the working memory $\vh_t$. 
	
The main innovation of the ARMIN cell is using 3 element-wise and soft gates ($i.e.\ \vg^h_t,\ \vg^r_t$, and $\vo^r_t$) to control the information flow for memory processing. \emph{Firstly}, by using $\vg^h_t$ and $\vg^r_t$ gates, the network can dependently filter out the irrelevant information for current time-step from $\vh_{t-1}$ and $\vr_t$ in an element-wise fashion, and keep the useful information for later information integration. This refined control can avoid noise and make the weight parameters easier to be trained. In extreme cases, if the read operation chooses a completely useless or wrong $\vr_t$, the network can shut down the $\vg^r_t$ gate in the first place; if the network needs to reset hidden state to a historical state, it can shut down $\vg^h_t$ gate and let $\vg^r_t$ open. As a result, these two gate can bring more fault-tolerance for training read operation and increase flexibility of the RNN hidden state transmission. \emph{Secondly}, by using the $\vo^r_t$ gate, we can $selectively$ output useful information from $\vr_t$ for later computations. \emph{Thirdly}, all gates in the ARMIN cell are element-wise and soft gates, which means we can smoothly apply existing RNN regularization techniques such as recurrent batchnorm and layer normalization to stabilize training process.

\subsection{Read Operation with Auto-addressing}
\label{read}

In this subsection, we introduce the auto-addressing mechanism. Specifically, the ARMIN maintains a memory matrix $\mM \in \R^{n_{mem}\times d_h}$, where the constant $n_{mem}$ denotes the number of memory slots. At each recurrence, the ARMIN chooses a historical state $r_t$ from memory according to the information in $\vx_t$ and $\vh_{t-1}$ ,which is formulated as follows:

\begin{equation}
\vs_t=\text{gumbel{\textrm{-}}softmax}(\mW_s[\ \vx_t\ ,\vh_{t-1}\ ]\ +\ \vb_s)\ ,
\end{equation}
\begin{equation}
\vr_t= \sum_{i=0}^{n-1}s_t(i)\mM(i,:)\ .
\end{equation}
where $\mW_s \in \R^{n_{mem} \times (d_i+d_h)}$, $\vb_s \in \R^{n_{mem}}$, $\vs_t$ is a one-hot vector sampled by gumbel{\textrm{-}}softmax function, $s_t(i)$ denotes the $i$-th element of $\vs_t$, $\mM(i,:)$ denotes the $i$-th row of $\mM$.

As opposed to previous fully-differentiable addressing mechanisms of recurrent MANNs such as NTM and TARDIS, the auto-addressing mechanism doesn't use the information from memory or any extra hand-crafted features to assist memory addressing, instead, it directly encodes the historical memory accessing information via the hidden state $\vh_{t-1}$, which way shows sufficient power for memory addressing in our empirical evaluations. The time and space complexities of auto-addressing are only proportional to the size of $d_i+d_h$, whereas the complexities of previous addressing mechanism are usually proportional to the size of entire memory. Furthermore, the simple form of the auto-addressing mechanism makes the gradient flow more straightforward, and leads to a faster training convergence speed and better performance, as is shown in our experiments.

\subsection{Write Operation}
\label{write}

After the reading and cell processing stages, the ARMIN writes the new hidden state $\vh_t$ to the memory $\mM$. Following previous memory networks, we simply overwrite $\vh_t$ to the memory slot where we just read out the $\vr_t$ (for conditions where $d_h$ is not equal to $d_r$, we first use a linear layer to transform $\vh_t$ from $d_h$ dimension to $d_r$ dimension); but at the initial time-steps, we write the hidden states to the empty memory slots, until all empty slots are filled with historical states. Our network shares some similar ideas with the TARDIS network, including the discrete and one-hot memory addressing and memory overwriting mechanism, but our network has a simpler addressing mechanism and more flexible cell computations. For a detailed theoretical comparison between TARDIS and ARMIN, please refer to our supplemental material\footnote{Please view it on \href{https://sites.google.com/view/armin-network}{ https://sites.google.com/view/armin-network.}}. We also compare them across our experiments to validate the efficiency of our network.

\section{Experiments}

We evaluate our model mainly on algorithmic tasks, pMNIST task and character-level language modelling task, and additionally the temporal action proposal task (please refer to the supplemental material). We compare our network with many previous MANNs and vanilla LSTM networks\footnote{We implement NTM based on \href{github.com/vlgiitr/ntm-pytorch}{github.com/vlgiitr/ntm-pytorch}, and  DNC, SAM based on \href{github.com/ixaxaar/pytorch-dnc}{github.com/ixaxaar/pytorch-dnc}. For SAB network, we only compare available results in their paper because the official code is not released.}. As our network is designed based on LSTM and the GRU has similar performance/cost with LSTM, we omit the comparison with the GRU. We also directly compare the auto-addressing against the TARDIS addressing mechanism, by replacing auto-addressing with the TARDIS addressing method in ARMIN. We refer to this network as AwTA (ARMIN with TARDIS-Addr) in our experiments. Please refer to the supplemental material for the implementation details of our experiments.

\subsection{Algorithmic Tasks}
\begin{figure}[!htb]
\centering
\begin{subfigure}[b]{.235\textwidth}
  \includegraphics[width=1.0\linewidth]{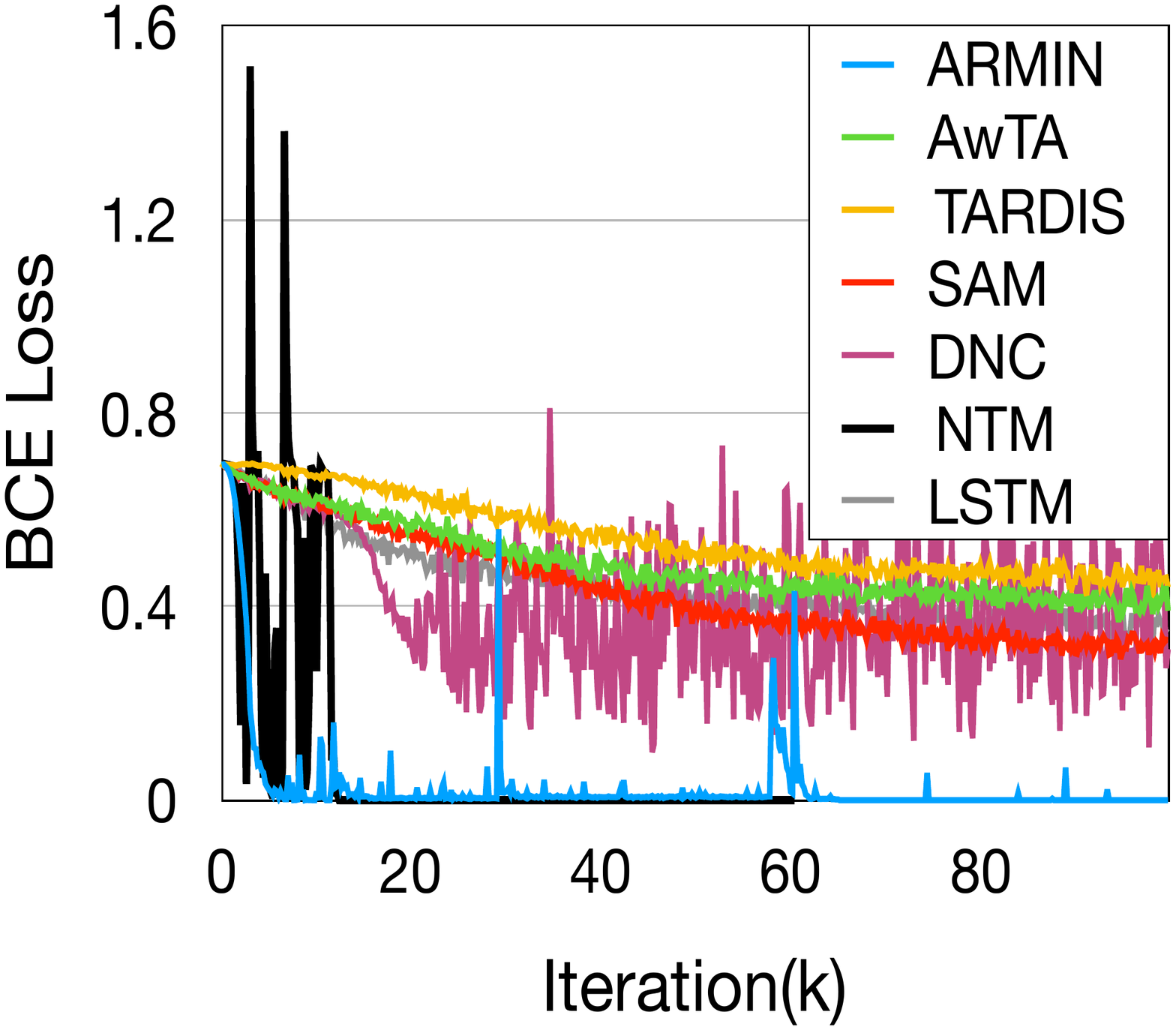}
 \vspace{-6\baselineskip}
  \caption{copy}
\end{subfigure}
\begin{subfigure}[b]{.235\textwidth}
  \includegraphics[width=1.0\linewidth]{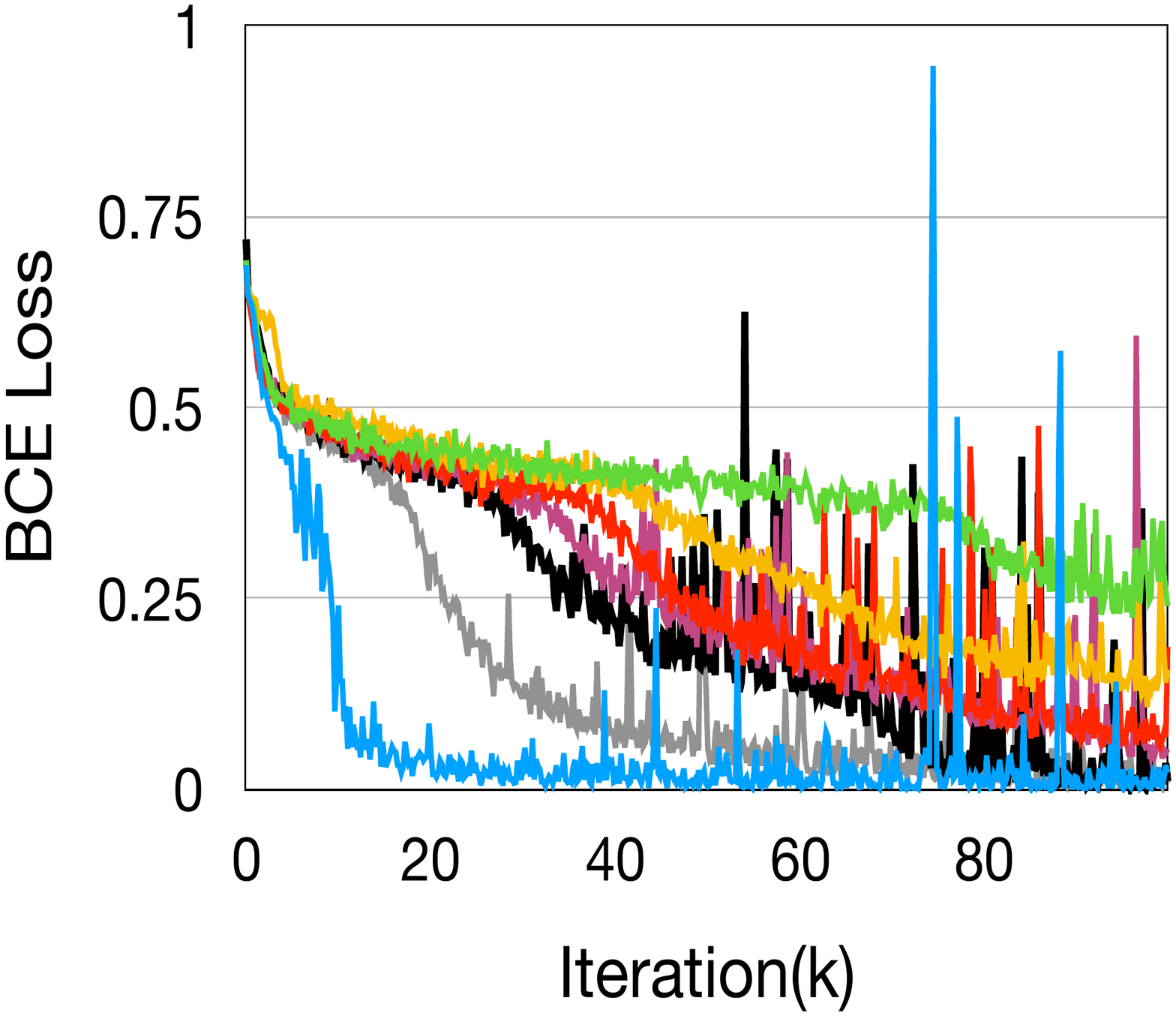}
  \vspace{-6\baselineskip}
  \caption{repeat copy}
\end{subfigure}
\begin{subfigure}[b]{.235\textwidth}
  \includegraphics[width=1.0\linewidth]{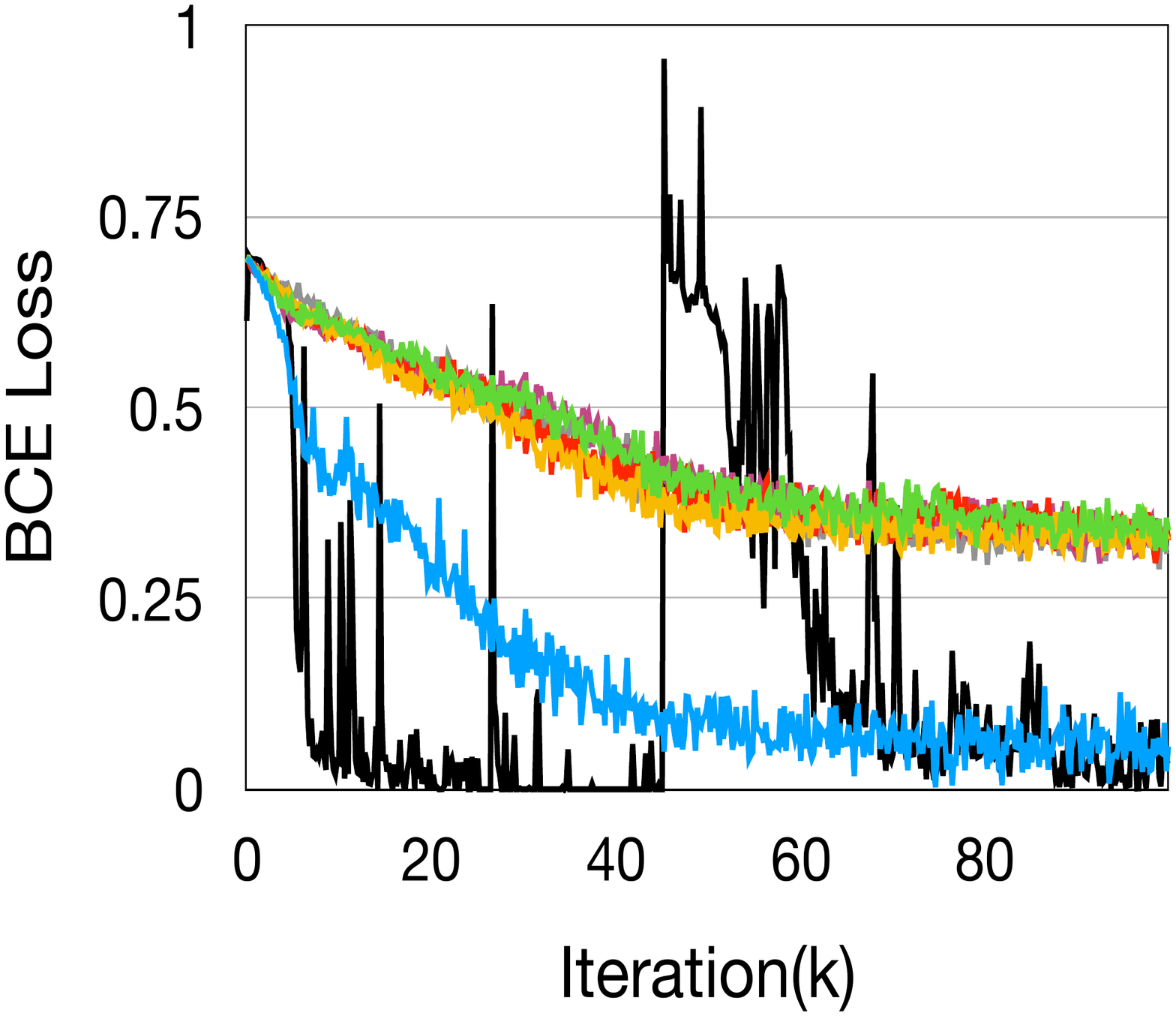}
  \vspace{-6\baselineskip}
  \caption{associative recall}
\end{subfigure}
 \vspace{-10\baselineskip}
\begin{subfigure}[b]{.235\textwidth}
  \includegraphics[width=1.0\linewidth]{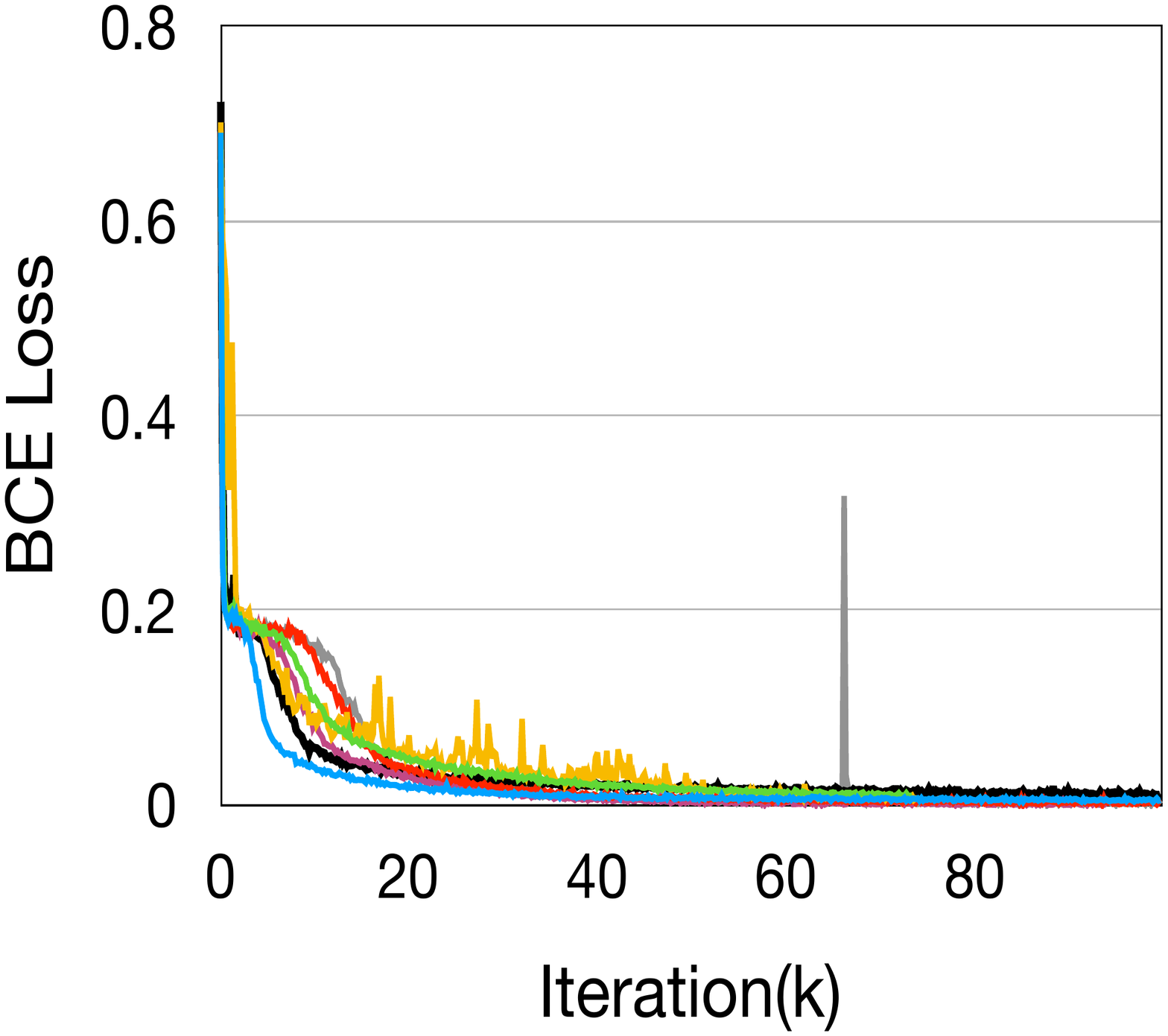}
 \vspace{-6\baselineskip}
  \caption{priority sort}
\end{subfigure}
 \vspace{10\baselineskip}
\caption{The curves of validation losses on algorithmic tasks.}
\label{figure algo}
\end{figure}

Along with the NTM, Graves $et\ al.$ \shortcite{graves2014neural} introduced a set of synthetic algorithm tasks to examine the efficiency of MANNs.  Here we use 4 out of 5 of these tasks (due to the page limit, we exclude the N-gram task that is not adopted in previous works such as SAM and TARDIS)
 to examine if our network can choose correct time-steps from the past and effectively make use of them: \emph{(a) copy}: copy a 6-bit binary sequence of length 1$\sim$50, \emph{(b) repeat copy}: copy a sequence of length 1$\sim$10 for 1$\sim$10 times, \emph{(c) associative recall} given a sequence of 2$\sim$6 (key,value) pairs and then a query with a key, return the associated value, \emph{(d) priority sort}: given a sequence of 40 (key, priority value) pairs, return the top 30 keys in descending order of priority. We keep the parameter counts of MANNs roughly the same,  and we use a strong LSTM baseline with about 4 times larger parameter count than MANNs.  Following Gulcehre \shortcite{gulcehre2017memory}, Campos \shortcite{campos2017skip}, in all tasks, we consider a task solved if the averaged binary cross-entropy loss of validation is at least two orders of magnitude below the initial loss which is around 0.70, $i.e.$ the validation loss (the validation set is generated randomly) converges to less than 0.01 (with less than $30\%$ in 10 consecutive validation of sharp losses that are higher than 0.01 ). In our evaluation, we are interested in if a model can successfully solve the task in 100k iterations and the iterations and elapsed training time till the model succeeds on the task. We run all experiments with the batch size of 1 under 2.8 GHz Intel Core i7 CPU, and report the wall clock time and the number of iterations for solved tasks. For models that fail to converge to less than 0.01 loss in 100k iterations, we report the average loss of the final 10 validations (denoted in underlines) and the elapsed training time for 100k iterations. The results are shown in Table \ref{table algo}. Please also see the supplemental material for the standard deviations of the losses of final 10 epochs of all models. The training curves for copy and priority sort tasks are shown in Figure \ref{figure algo}.

\newcommand{\tabincell}[2]{\begin{tabular}{@{}#1@{}}#2\end{tabular}}
\begin{table*}[!t]
\begin{center}
\begin{tabular}{c|r|r|r|r|r|r|r|r|r|r|r}
\hline
  &\tabincell{c}{\bf Hidden \\ \bf Size}&\tabincell{c}{\bf Mem. \\ \bf Size}&\tabincell{c}{\bf Param. \\ \bf Count} & \multicolumn{2}{c}{\bf Copy} & \multicolumn{2}{|c}{\tabincell{c}{\bf Repeat \\ \bf Copy}} & \multicolumn{2}{|c}{\tabincell{c}{\bf Associative \\ \bf Recall}}  & \multicolumn{2}{|c}{\tabincell{c}{\bf Priority \\ \bf Sort}}  \\ \cline{2-12}
     & & &  & iter. & time & iter. & time & iter. & time & iter. & time  \\
\hline
LSTM &300 &-- &376k  & \underline{0.359}& 66  & \underline{0.016}& 45 & \underline{0.325}  & 34 & 37.0  & 32\\
NTM  &120 &128$\times$20 &88k  & 12.4 & 32  & \underline{0.014} & 171 & 19.6 & 22& \underline{0.012} & 360 \\
DNC  &120 &128$\times$20 &104k  & \underline{0.429}& 348 & \underline{0.053} & 208& \underline{0.338} & 145 & 34.4  & 160 \\
SAM  &120 &128$\times$20 & 109k  & \underline{0.321} & 145 & \underline{0.064} & 87& \underline{0.326} & 64  & 78  & 40\\
TARDIS  & 120&50$\times$32 &90k  & \underline{0.451} & 157 & \underline{0.166}& 97& \underline{0.330} & 66  & 62.6 & 133 \\
AwTA &100 &50$\times$32 &91k & \underline{0.410}& 152  & \underline{0.297} & 95 & \underline{0.336} & 64 & 71.4  & 151\\
ARMIN  &100 &50$\times$32 &90k  & 7.6& 6& 67.8  & 36  & \underline{0.052}& 33 & 33.4  & 37\\
\hline
\end{tabular}
\caption{The average iterations(k) and elapsed time(min) of different networks till task solved in the given 100k iterations. The wall clock training time for 100k iterations and the average losses of final 10 validations (denoted in underlines) are shown for the unsolved tasks.}
\label{table algo}
\end{center}
\end{table*}

The results show that the ARMIN can solve 3 out of 4 tasks with a fast converge speed in terms of both wall clock time and iteration numbers. By comparing the ARMIN with NTM, we observe that the NTM is able to almost solve all of the tasks, but with a much slower training speed compared to the ARMIN, for example, the training time of NTM is 5 times larger than ARMIN to solve the copy task. In fact, the training speed of ARMIN for each iteration is about $3\sim4$ times faster than the NTM's speed. We observe that the TARDIS and AwTA fail to solve 3 out of 4 tasks in the given 100k iterations. By comparing them with the ARMIN, \emph{we confirm the efficiency of the auto-addressing of ARMIN}. The superiority of ARMIN in these tasks can be ascribed to the auto-addressing that favors a straightforward gradient propagation, and makes itself easier to learn, as in sec \ref{read}. We note that in the same loss range, ARMIN curves are smoother than other MANNs. The peaks at low loss are because part of the correct output array shifts left or right by one time-step, as explained in Graves \shortcite{graves2014neural}.

\begin{table*}[htb]
\centering
\begin{tabular}{crrrrrr}
\toprule
\multicolumn{1}{c}{\bf Model}  &\tabincell{c}{\bf Test \\ \bf Acc.(\%)}  &\tabincell{c}{\bf Hidden \\ \bf Size} &\tabincell{c}{\bf Memory \\ \bf Size}&\tabincell{c}{\bf Param. \\ \bf Count} &\tabincell{c}{\bf GPU \\ \bf memory(GB)} &\tabincell{c}{\bf Time \\ \bf (min/epoch)} \\
\midrule
LSTM &92.7 &128 &-- & 69k &1.687 & 7\\
NTM  &92.5 &100 &28$\times$28 & 68k &2.912 & 40\\
DNC  &94.3 &100 &28$\times$28 & 70k &7.908 & 48\\
SAM  &NaN &100 &28$\times$28 & 72k & -- & --\\
SAB  &94.2 &-- &-- & -- & -- & --\\
TARDIS &94.1  &100 &28$\times$28 & 74k &6.919 & 24\\
AwTA &94.1  &100 &28$\times$28 & 81k &7.884 & 23\\
ARMIN &94.3  &100 &28$\times$28 & 81k & 2.713 & 15\\
\bottomrule
\end{tabular}
\caption{Test accuracy and memory/time consumption during training on pMNIST task.}
\label{table pmnist}
\end{table*}

\subsection{Permuted Pixel-by-pixel MNIST Classification}
The Permuted pixel-by-pixel MNIST classification ($p$MNIST) task consists of predicting the handwritten digits of images on the MNIST dataset after being given its 784 pixels as a sequence permuted in a fixed random order. The model outputs its prediction of the handwritten digit at the last time step, so that it must integrate and remember observed information from previous pixels. We use this task to validate if our network can better store and integrate the information of previously observed pixels when the local structure of the image is destroyed by the pixel permutation. We run the experiments using double-precision floating point calculation under a TESLA K80 GPU and report the test set accuracy along with the time and memory consumption of the training stage.  

The results are shown in Table \ref{table pmnist}. \emph{ARMIN and DNC outperform other networks in classification accuracy, but the ARMIN saves 65.6$\%$ GPU memory and 62.5$\%$ training time compared to DNC}. We observe that ARMIN outperforms LSTM in terms of accuracy but not in terms of memory and time consumption. There indeed exists a performance/cost trade-off when comparing LSTM and ARMIN in sequence classification tasks such as pMNIST, because the memory addressing learning (which we assume is relatively hard compared with learning the parameters of LSTM) can only receive gradients from the classification label at the last time-step. However, this won't happen in sequence tagging tasks such as language modelling, as we will show in section \ref{towards}.

\subsection{Character-level Language Modelling}
\label{task lm}
The character-level language modelling task consists of predicting the probability distribution of the next character given all the previous ones. We benchmark our network over the Penn Treebank and Hutter Prize Wikipedia (also known as $enwik8$) datasets. In this experiment, we also compare our network with some state-of-the-art RNN variants on these datasets, such as HM-LSTM \cite{chung2016hierarchical}, HyperLSTM \cite{ha2016hypernetworks}, NASCell \cite{zoph2016neural}, IndRNN \cite{li2018independently}, HyperRHN \cite{suarez2017language} and FS-LSTM \cite{mujika2017fast}. 

We ensure similar parameter counts and the same hyperparameters for all MANNs. Layer normalization (LN) \cite{ba2016layer} and zoneout \cite{krueger2016zoneout} are applied for the MANNs to avoid overfitting. Following prior works, we apply truncated backpropagation through time(TBPTT) \cite{rumelhart1986learning,elman1990finding} to approximate the gradients: at each iteration, the network predict the next 150 characters, and the hidden state $\vh_t$ and memory state $\mM$ are passed to the next iteration. The gradients are truncated between different iterations.

The results are shown in table \ref{table char}. On Penn Treebank dataset, our best performing network achieves competitive 1.198 BPC on Penn Treebank dataset, which is the best single cell performance that we are aware of. \emph{By comparing TARDIS and AwTA, we observe about 4 points of improvement, which shows the efficiency of the ARMIN cell. By comparing AwTA and ARMIN, we observe a further improvement of around 2 points, which shows the efficiency of the auto-addressing mechanism}. We also find that the same architecture of NTM and DNC that perform well in algorithmic tasks or $p$MNIST task fail to converge to a lower BPC than vanilla LSTM, which in turn shows the \emph{generality} of ARMIN network.

The results on Penn Treebank show our single ARMIN cell learns better representations than many hierarchical RNN structures, such as the HM-LSTM, 2-Layer HyperLSTM and 21 layer IndRNN. Our network is outperformed by HyperRHN and FS-LSTM which are both multi-scale and deep transition RNNs and are state-of-the-art RNNs on this dataset. By ``better representations'', we refer to the concatenation of the gated contents from $\vh_t$ and $\vr_t$ as in equation \ref{eq13}. If we remove the gated contents of $\vr_t$ from $\vo_t$, the ARMIN undergoes a BPC performance drop from 1.198 to 1.220, which is still better than the best performing BPC of 1.24 of the LSTM. We believe the rest of the performance gain comes from the recurrent memory integration of the ARMIN cell, which also implicitly favors the function of deep transition, as is shown by the success of the deep transition RNNs on this task.

For more ablation study on the Penn Treebank dataset regarding the auto-addressing mechanism and the ARMIN cell, please refer to the supplemental material.

\begin{table}[H]
\begin{center}
\begin{tabular}{lrrrr}
\toprule
  &\multicolumn{2}{c}{\bf PTB} &\multicolumn{2}{c}{\bf enwik8}\\ 
 \multicolumn{1}{c}{\bf Model} &BPC &\tabincell{c}{Params} &BPC &\tabincell{c}{Params}   \\
\midrule
LSTM  &1.36 &-- &1.45 &--\\
LSTM+Zoneout  &1.27 &-- &-- &--\\
LSTM+LN  &1.267 &4.26M &1.39 &--\\
LSTM+LN+Zoneout &1.24 &4.79M &1.37 &--\\
[-1.5ex]\\ \hline \\[-1.5ex]
HM-LSTM+LN    &1.24 &-- &1.32 &35M\\
HyperLSTM+LN     &1.219 &14.41M &1.340 &26.5M\\
NASCell        &1.214  &16.28M&-- &--\\
IndRNN (21 layers)          &1.21  &--&-- &--\\
HyperRHN  &1.19  &15.5M&-- &--\\
\tabincell{c}{FS-LSTM+L- \\ N+Zoneout}     &1.190  &7.2M &1.277 &27M\\
[-1.5ex]\\ \hline \\[-1.5ex]
NTM(800 units)  &1.535 &8.28M &-- &--\\
DNC(800 units) &1.390 &8.20M &-- &--\\
SAM(800 units) &NaN &9.48M &-- &--\\
SAB(paper) &1.37 &9.48M &-- &--\\
TARDIS(paper)  &1.25 &-- &-- &--\\
TARDIS(1000 units) &1.268 &9.2M &-- &--\\
\tabincell{c}{AwTA(800 units)} &1.223 &10.2M &-- &--\\
\midrule
ARMIN (500 units)      &1.236  &4.03M&-- &--\\
ARMIN (800 units)    &1.198  &9.80M&-- &-- \\
\tabincell{c}{ARMIN \\ (800 units, 2 layer)}    &-- &--  &1.331  &21.6M \\
\bottomrule

\end{tabular}

\caption{Bits-per-character on Penn Treebank and enwik8 test set. The lower is the better.}
\label{table char}
\end{center}
\end{table}

The result on enwik8 demonstrates a simple 2-layer ARMIN can achieve competitive BPC performance of 1.33, with less parameter count compared to the HyperLSTM and HM-LSTM. By constructing deeper ARMIN network or even combining with other multi-scale and hierarchical RNN architectures, we believe the performance can be further improved.

\section{Towards a More Light-weight Recurrent Memory Network}
\label{towards}

\begin{figure}[!htbp]
\setlength{\abovecaptionskip}{.235cm}
\setlength{\belowcaptionskip}{-0.1cm}
\begin{subfigure}[b]{.47\textwidth}
  \includegraphics[width=1.1\linewidth]{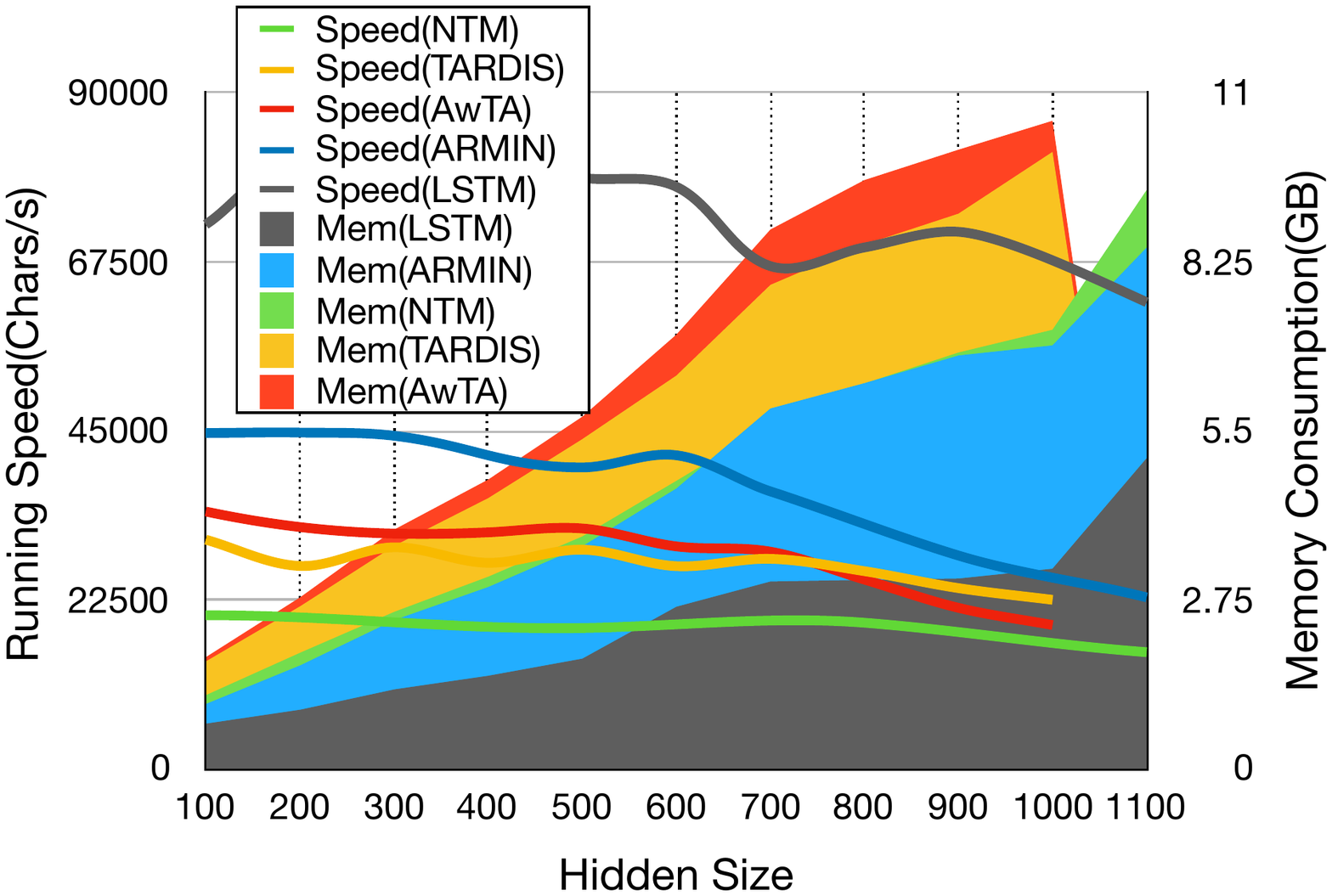}
 \vspace{-17\baselineskip}

  \caption{Training (batch size=128)}
\end{subfigure}
\begin{subfigure}[b]{.47\textwidth}
  \includegraphics[width=1.1\linewidth]{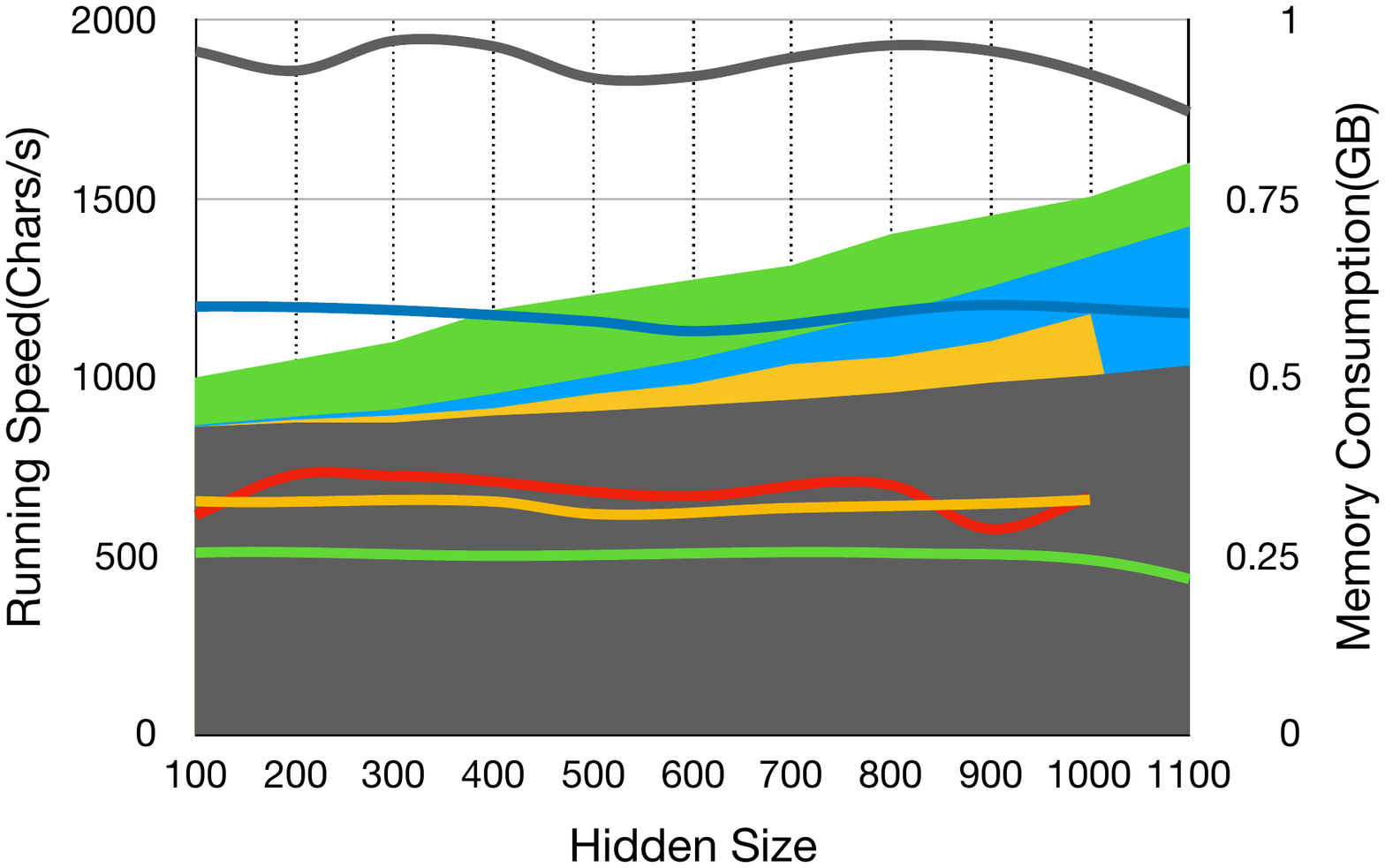}
  \vspace{-17.8\baselineskip}
  \caption{Inference (batch size=1)}
\end{subfigure}

\caption{The running speed and memory consumption at the training and inference stages ($n_{mem}=20$). The solid blocks denote the memory consumption (GB) and the curves denote the running speed (characters/s). {\bf(a)} shows the training stage, and {\bf(b)} shows the inference stage. DNC and SAM is not shown in the graph, becuase 1.The time and memory consumption of DNC is worse than NTM. 2.SAM doesn't converge in our pMNIST and language modelling tasks. Note that AwTA has almost the same inference memory consumption with ARMIN.}
\label{figure light}
\end{figure}

 \paragraph{Thorough Comparison of MANNs.} In previous sections, we have shown in some of the algorithmic tasks that the ARMIN can be trained $3\sim4$ times faster than the NTM. In the next, we conduct a more thorough comparison among the vanilla LSTM and memory networks under different hidden units. Without the loss of generality, we do the benchmarks on Penn Treebank dataset and reuse the experimental setup as in section \ref{task lm}. We keep all memory matrices the same size, $i.e.\ 20 \times d_h $. We run the experiment using single-precision under a Titan XP GPU that has 12GB memory space. The results are depicted in figure \ref{figure light}. From the results we conclude that the ARMIN consistently outperforms other memory networks shown in the graph in terms of running speed both at training and inference stages, and the main contribution to this comes from the simple auto-addressing mechanism of the ARMIN. Moreover, at inference stage, \emph{we can replace the memory matrix with a list of discrete memory slots, and update memory by simply replacing the old hidden states with the new ones, furthermore, we can replace the gumbel-softmax function with argmax}. Using these methods, ARMIN's inference speed obtains significant improvement than in training stage. At inference stage, we observe TARDIS has smaller memory consumption than ARMIN, this is due to the fact that TARDIS has smaller parameter counts under the same hidden size. Actually, TARDIS and ARMIN have roughly the same inference memory consumption under the same parameter counts. 
 
 \begin{table}[t]
        \centering
        \begin{tabular}{c|c|c|c|c} 
        \toprule
        Model&  \multicolumn{2}{c|}{\bf LSTM} & \multicolumn{2}{c}{\bf ARMIN}\\ \cline{2-5}
		\hline
		Setup& 1 &2 &1 &2\\
		\hline
          Hidden size & 1k & 1k & 500 & 550 \\
          $n_{param}$(M) & 4.79 & 4.79 & 4.02 & 4.81 \\
          $n_{mem}$ & -- & -- & 5 & 10 \\
          $T_{trunc}$  & 100 & 150 & 50 & 50\\       
          batch size & 128& 128  & 384 & 300\\
          Memory(GB)  & 2.36 & 3.27 & 3.49 &3.56\\
          \tabincell{c}{Speed \\ (chars/s)}  & 71k & 70k & 98k & 75k\\
          BPC  & 1.27 & 1.24 & 1.238 & 1.223 \\
\bottomrule
 \end{tabular}
 \caption{Comparison for performance and cost of different setups on Penn Treebank dataset.}
        \label{table light}
\end{table}

\paragraph{Accelerating ARMIN by Short-length TBPTT.} TBPTT is widely used in long sequential tasks to decrease training memory consumption and speed up the training process. However, it's obvious that a trade-off exists when applying TBPTT: the shorter the truncate length is, the worse the RNN performs. The subsequent sequences can only receive a single hidden state without gradient backpropagation enabled from their previous sequences, and thus, the RNNs can't access historical hidden states via backpropagations and effectively learn long-term dependencies. To alleviate this issue, we explore combining external memory with short-length TBPTT to allow the truncated sequences to receive more information from history while keeping the truncate length short, and thus, can accelerate the training process using large batch size while keeping good performance and similar memory consumption. Specifically, when we shorten the TBPTT length from 150 to 50 in the language-modelling task in section \ref{task lm}, we find the LSTM undergoes a big BPC performance drop from 1.24 to 1.39, but the ARMIN only has minor performance drop from 1.198 to 1.223, which shows the efficiency of combining the external memory with TBPTT. We further conduct an experiment to compare the ARMIN and LSTM under different setups. The hyperparameter setups and results are shown in Table \ref{table light}. By comparing ARMIN setup 1 with LSTM setup 2, we observe that with only $6.7\%$ more memory consumption, we \emph{obtain $40\%$ training speed gain} while keeping a slightly better BPC performance and $16\%$ less parameter count; by comparing ARMIN setup 2 with LSTM setup 2, we observe that with only $8.8\%$ more memory consumption, we obtain about 1.7 points of BPC performance gain and $7.1\%$ training speed gain under similar parameter count. The results imply that on the long sequential tasks, we can use ARMIN to boost training speed or achieve better performance depending on the specific cases.

\section{Conclusion}
In this paper, we introduce the ARMIN, a light-weight MANN with a novel ARMIN cell. The ARMIN incorporates an efficient external memory with the light-weight auto-addressing mechanism. We demonstrate competitive performance of ARMIN in various tasks, and the generality of our model. It's observed that our network is robust to short-length TBPTT which enables using large batch size to speed up the training while keeping good performance.

\section*{Acknowledgements}
This work was supported in part by the Project of Shenzhen Municipal Science and Technology Program (No. JCYJ20170818141146428), in part by the National Engineering Laboratory for Video Technology-Shenzhen Division, in part by Shenzhen Key Laboratory for Intelligent Multimedia and Virtual Reality (No. ZDSYS201703031405467), and in part by National Natural Science Foundation of China and Guangdong Province Scientific Research on Big Data (No. U1611461). We appreciate our anonymous reviewers for their valuable comments. In addition, we would like to thank Jerry for English language editing.

\bibliographystyle{named}
\bibliography{ijcai19}

\end{document}

%% file: math_commands.tex
\usepackage{amsmath,amsfonts,bm}

\def\eqref#1{equation~\ref{#1}}

\def\1{\bm{1}}

\def\vb{{\bm{b}}}

\def\vf{{\bm{f}}}
\def\vg{{\bm{g}}}
\def\vh{{\bm{h}}}
\def\vi{{\bm{i}}}

\def\vo{{\bm{o}}}
\def\vp{{\bm{p}}}

\def\vr{{\bm{r}}}
\def\vs{{\bm{s}}}

\def\vx{{\bm{x}}}
\def\vy{{\bm{y}}}

\def\mM{{\bm{M}}}

\def\mW{{\bm{W}}}

\DeclareMathAlphabet{\mathsfit}{\encodingdefault}{\sfdefault}{m}{sl}
\SetMathAlphabet{\mathsfit}{bold}{\encodingdefault}{\sfdefault}{bx}{n}

\newcommand{\R}{\mathbb{R}}

